\documentclass[11pt]{article}

\usepackage[preprint]{acl}
\usepackage{times}
\usepackage{latexsym}
\usepackage[T1]{fontenc}
\usepackage[utf8]{inputenc}
\usepackage{microtype}
\usepackage{inconsolata}
\usepackage{graphicx}
\usepackage{booktabs}
\usepackage{multirow}
\usepackage{amsmath}
\usepackage{amssymb}
\usepackage{amsthm}
\usepackage{stfloats}
\usepackage{tabularx}
\usepackage{multicol}
\usepackage{xcolor}
\usepackage{subcaption}
\usepackage{array}
\usepackage{fontawesome5}
\usepackage{siunitx}
\usepackage{url}
\usepackage{tcolorbox}
\tcbuselibrary{skins, breakable}
\usepackage{xcolor}
\newcommand{\gamedai}{\textsc{GamED.AI}}

\title{\gamedai{}: A Hierarchical Multi-Agent Framework for Automated Educational Game Generation}

\author{
 Shiven Agarwal \thanks{contributed equally} \quad
Yash Shah  \footnotemark[1] \quad
Ashish Raj Shekhar\footnotemark[1] \quad
 Priyanuj Bordoloi\\
 \quad
\textbf{Vivek Gupta} \\
Arizona State University \\
\faGlobe~\href{https://shivena99.github.io/GamED-AI/acl-demo/}{Project Page} \quad
\faPlayCircle~\href{https://shivena99.github.io/GamED-AI/acl-demo/library/}{Demo} \quad
\faVideo~\href{https://youtu.be/3aMklNAMysM}{Video} \quad
\faGithub~\href{https://github.com/ShivenA99/GamED-AI}{Code} \\
\texttt{\{ sagar147, yshah124, ashekha9,  pbordolo, vgupt140\}@asu.edu}
}
\begin{document}
\maketitle

\begin{abstract}
We introduce \gamedai{}, a hierarchical multi-agent framework that
transforms instructor-provided questions into fully playable,
pedagogically grounded educational games validated through formal
mechanic contracts. Built on phase-based LangGraph sub-graphs,
deterministic Quality Gates, and structured Pydantic schemas,
\gamedai{} supports two template families encompassing 15 interaction
mechanics across spatial reasoning, procedural execution, and
higher-order Bloom's Taxonomy objectives. Evaluated on 200 questions spanning five subject domains, the system achieves a 90\% validation pass rate against internal
FOL-based structural validators (an architectural compliance metric,
not an independent pedagogical benchmark), 98.3\% schema compliance,
and 73\% token reduction over ReAct agents (${\sim}$73,500
$\rightarrow$ ${\sim}$19,900 tokens/game) at \$0.46 per game. Within this model
configuration, these results suggest that phase-bounded architectural
structure correlates more strongly with alignment quality than
prompting strategy alone. Our
demonstration lets attendees generate Bloom's-aligned games from
natural language in under 60 seconds, inspect Quality Gate outputs
at each pipeline phase, and browse a curated library of 50 games
spanning all 15 mechanic types.
\end{abstract}
\section{Introduction}

Large Language Models now resolve 50--64\% of real-world engineering
tasks \citep{jimenez2024} and achieve high Pass@1 rates on function-level benchmarks
\citep{chen2021,openai2023}, yet their effectiveness in
producing \textit{pedagogically valid} educational content remains
limited---particularly where Bloom's Taxonomy alignment, mechanic
contract enforcement, and structured competency evidence are required
\citep{mislevy2003,shute2013}.

This gap matters, as game-based assessments achieve meta-analytic effect
sizes of $g = 0.49$ on cognitive outcomes \citep{sailer2020},
$g = 0.78$ on academic performance \citep{zeng2024}, and $d = 0.29$
on learning \citep{wouters2013}---yet a single publication-ready game
exceeds \$10,000 to produce \citep{chapman2010}, with mechanics
routinely decoupled from objectives on existing platforms
\citep{wang2020}. General-purpose agentic tools fall short on two
grounds: missing Bloom's-level targeting, and self-correction loops
that inflate tokens and accumulate errors
\citep{yao2023,brittle_react2024}---yielding games syntactically
correct but semantically wrong, e.g., testing \textit{recall} when
the objective requires \textit{analysis} \citep{mislevy2003,ji2023}.

We introduce \gamedai{}, a hierarchical multi-agent framework that
transforms instructor-provided questions into Bloom's-aligned
educational games validated through formal mechanic contracts. Built
on a LangGraph DAG with phase-specific sub-graphs, deterministic
Quality Gates, and typed Pydantic schemas, \gamedai{} reduces the error propagation that makes prior agentic architectures impractical for structured content generation
\citep{brittle_react2024,yao2023}. The system generates validated games in under 60 seconds at \$0.46
per game---achieving 73\% token reduction over ReAct agents
\citep{yao2023} and 90\% validation pass rate across 200 test
questions covering all 15 mechanics. Our contributions:

\begin{itemize}
  \item To our knowledge, the first hierarchical multi-agent framework
  for educational game generation, with 15 interaction mechanics and
  Bloom's alignment contracts enforced before generation. 
  \item 90\% validation pass rate and 73\% token reduction over ReAct
  agents, outperforming Claude Code on Bloom's alignment under all
  four prompting conditions.
  \item A live demo enabling real-time game generation, pipeline
  observability, and a browsable library of 50 curated games spanning
  all 15 mechanics are open-sourced.
\end{itemize}
\section{Related Work}
Bloom's Taxonomy \citep{anderson2001} and Evidence-Centered Design
\citep{mislevy2003} require game mechanics to constitute valid
competency evidence; the LM-GM framework \citep{arnab2015} provides
the mechanic-mapping heuristic informing our Bloom's constraint table
(Appendix~\ref{app:blooms_mapping}). Gamification succeeds when
mechanics match learning goals \citep{sailer2020}, fails when
decorative \citep{hamari2014}; formative feedback design
\citep{shute2008} informs per-element validation at QG3.

Satisfying these constraints programmatically requires architectures
that prevent error propagation. MetaGPT \citep{hong2023} and AutoGen
\citep{wu2023} use role-bounded schemas; ReAct \citep{yao2023} adds
self-correction but produces token inflation
\citep{brittle_react2024}; flow engineering \citep{ridnik2024} and
constrained decoding \citep{willard2023} make invalid states
structurally unreachable. Widely adopted platforms (Kahoot, H5P) lack
objective alignment \citep{wang2020}; GameGPT
\citep{chen2023gamegpt} targets speed without Bloom's alignment.
\gamedai{} integrates Bloom's alignment, FOL-based validation, and a
modular game engine in a single open-source framework.

\section{\gamedai{}: Bridging Pedagogical Intent and Generative
Execution}

\begin{figure*}[t]
    \centering
    \includegraphics[width=\textwidth]{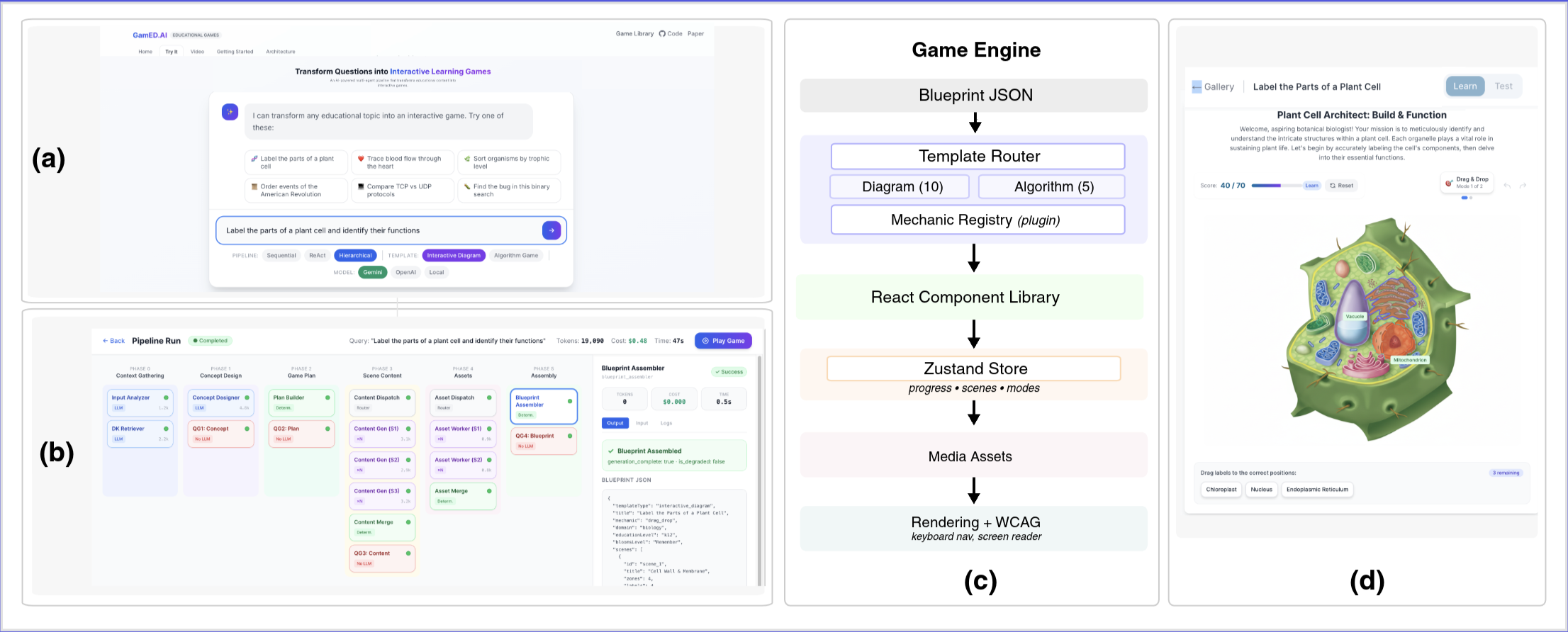}
    \caption{\textbf{End-to-end system demonstration.}
    \textbf{(a)}~Instructor enters a natural language question.
    \textbf{(b)}~DAG pipeline with real-time observability (per-agent
    traces, token/cost analytics, Quality Gate decisions).
    \textbf{(c)}~Modular game engine: Blueprint JSON is routed via
    Template Router and Mechanic Registry to self-contained React
    components with shared Zustand state and WCAG rendering.
    \textbf{(d)}~Generated Interactive Diagram Game (drag-and-drop)
    for ``Label the Parts of a Plant Cell.''}
    \label{fig:system_flow}
\end{figure*}

\gamedai{} accepts a natural language question or topic---with optional context (subject domain, target audience, difficulty level)---and produces a fully playable game, a structured alignment report, and a validation certificate confirming that all mechanic contracts are satisfied. As demonstrated in Figure~\ref{fig:system_flow}, the end-to-end pipeline transforms instructor input into an interactive, pedagogically grounded educational game in under 60 seconds. Four design principles govern all architectural decisions:
\begin{itemize}
  \item \textbf{Pedagogical primacy:} Every game is bound to a Bloom's level before generation; mechanic selection follows learning objectives \citep{anderson2001,mislevy2003}.
  \item \textbf{Deterministic validation:} Every generative step is gated by a non-stochastic validator; LLM outputs are proposals subject to structural verification \citep{ji2023}.
  \item \textbf{Structure over retry:} Typed schemas and phase boundaries prevent errors rather than catching them downstream \citep{willard2023}.
  \item \textbf{Modularity:} New templates are registered via contract definition without modifying orchestration \citep{hong2023,wu2023}.
\end{itemize}

\subsection{Architectural Evolution}

The current DAG architecture supersedes two prior designs: a
\textbf{Sequential Pipeline} (56.7\% VPR,
${\sim}$45,200 tokens/game) and a \textbf{ReAct Agent} system
(72.5\% VPR, ${\sim}$73,500 tokens/game). A key design target across
all iterations was sub-\$0.50 per-game cost, derived from the
Chapman Alliance~\citeyearpar{chapman2010} benchmark for scalable
content authoring; neither prior design met this threshold. Quantitative
comparison of all three architectures is shown in
Figure~\ref{fig:arch_comparison}. The detailed pipeline diagrams of
the earlier versions are available in Appendix~\ref{app:evolution}.

\subsection{\gamedai{} (DAG) Architecture}

The current architecture emerged from the observation that prior designs conflated generation and validation into the same cognitive
loop. The DAG separates them into six deterministic phases, each
bounded by a Quality Gate.

\subsubsection{System Architecture}
\label{sec:sysarch}
\begin{figure*}[htbp]
    \centering
    \includegraphics[width=\textwidth]{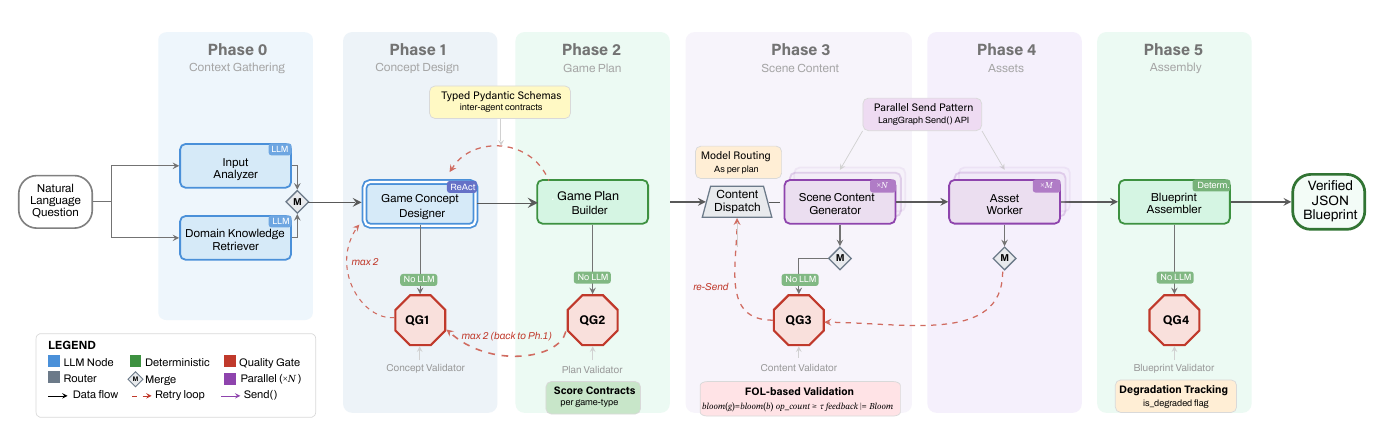}
    \caption{\textbf{\gamedai{} DAG architecture.} The pipeline
    comprises six phases: \textbf{Context Gathering},
    \textbf{Concept Design}, \textbf{Game Plan},
    \textbf{Scene Content}, \textbf{Assets}, and \textbf{Assembly}.
    Each phase operates as an independent sub-graph bounded by a
    deterministic Quality Gate (QG1--QG4) enforced without LLM
    inference. QG3 applies FOL-based Bloom's alignment predicates:
    $\text{bloom}(g) = \text{bloom}(b)$,
    $\text{op\_count} \geq \tau$, and
    $\text{feedback} \models \text{Bloom}$. Inter-agent contracts are
    governed by typed Pydantic schemas; Phases~3--4 use parallel
    \texttt{Send()} patterns to dispatch $N$ scene content workers
    and $M$ asset workers. Dashed edges denote bounded retry loops
    (max 1--2).}
    \label{fig:pipeline}
\end{figure*}

The system is a hierarchical DAG in LangGraph with six
phases---\textbf{Context Gathering}, \textbf{Concept Design},
\textbf{Game Plan}, \textbf{Scene Content}, \textbf{Assets}, and
\textbf{Assembly}---each an independent sub-graph with typed I/O and a
Quality Gate at its boundary (QG1--QG4;
Figure~\ref{fig:pipeline}). No agent in phase $N$ receives input
from phase $N{+}1$; no gate can be bypassed; invalid states cannot
propagate---a structural guarantee of the DAG topology
\citep{ridnik2024,hong2023}.

\subsubsection{Phase 0: Context Gathering}
The pipeline opens with two parallel LLM nodes: an
\textbf{Input Analyzer} that parses subject domain, target audience,
and difficulty level from the natural language question, and a
\textbf{Domain Knowledge Retriever} that grounds generation in curated
sources (textbooks, curriculum standards, domain ontologies). Outputs
are merged before Phase~1, ensuring concept design operates on
verified domain context rather than open-ended generation
\citep{mislevy2003}.

\subsubsection{Game Template Architecture}
The generative surface comprises \textbf{two template families} with
\textbf{15 interaction mechanics}.
\textbf{Interactive Diagram Games} (10 mechanics: drag-and-drop,
click-to-identify, trace-path, description matching, sequencing,
sorting, memory match, branching scenario, compare/contrast,
hierarchical) operate on spatial and relational content targeting
visual and conceptual reasoning \citep{mayer2009,sweller1988}.
\textbf{Interactive Algorithm Games} (5 mechanics: state tracer, bug
hunter, algorithm builder, complexity analyzer, constraint puzzle)
operate on procedural content targeting \textit{applying},
\textit{analyzing}, and \textit{creating} objectives, grounded in
algorithm visualization research
\citep{hundhausen2002,naps2002,anderson2001} and debugging-first
pedagogy \citep{lee2014,koedinger2006}. Together, these support a
library of \textbf{50 curated games} across five domains; the full
Bloom's-to-mechanic mapping is in
Appendix~\ref{app:blooms_mapping}.

\subsubsection{Scene and Mechanic Composition}
Templates span three structural configurations resolved automatically
from Bloom's level and content complexity:

\textbf{Single-scene, single-mechanic}---one interaction type, one
content context; covers ${\sim}$35\% of the library.

\textbf{Single-scene, multi-mechanic}---2--3 interaction types within
one content frame, validated through a state machine ensuring
compatible I/O schemas; covers ${\sim}$40\% \citep{sweller1988}.

\textbf{Multi-scene, multi-mechanic}---2--4 causally connected scenes
with monotonically increasing Bloom's levels, bounded by cognitive
load constraints ($\leq$4 scenes, $\leq$3 mechanics/scene); covers
${\sim}$25\% \citep{sweller1988}. In the 200-question evaluation,
these proportions held at 34\%, 41\%, and 25\% respectively across
the stratified question set; the remaining $<$1\% were re-routed to
single-scene configurations by QG1 on Bloom's under-specification.

\subsubsection{Mechanic Contracts and Blueprint Generation}
Template selection is a \textbf{constrained inference} in Phase~1:
the Game Concept Designer (ReAct) resolves input against a
Bloom's-to-mechanic constraint table encoding valid competency
evidence \citep{mislevy2003}. The result is a \textbf{Game
Blueprint}---a validated Pydantic document specifying learning
objective, Bloom's level, template, and mechanic contract---produced
before content generation begins and certified by QG1. Each contract
defines the interaction primitive, content types, valid Bloom's range,
and completion conditions, enforcing pedagogical alignment as a
structural constraint \citep{anderson2001,shute2013}. QG2
subsequently validates the full Game Plan against Score Contracts per
game type before any content generation begins.

\subsubsection{Generation and Assembly}
\label{sec:bloomqg3}

\paragraph{Parallel Generation.}
Phase~2 (Game Plan) produces a validated blueprint that gates entry
into the parallel generation stages. Phase~3 dispatches $N$ parallel
\textbf{Scene Content Generators} via LangGraph \texttt{Send()}
patterns; Phase~4 dispatches $M$ parallel \textbf{Asset Workers}
producing visual assets (SVG diagrams or text-synthesised visuals),
instructional text (directions, hints, per-element feedback), and
interaction specifications (drag targets, click regions, sequence
orders).

\paragraph{FOL-Based Content Validation.}
QG3 validates all content against FOL-based Bloom's alignment
predicates---$\text{bloom}(g) = \text{bloom}(b)$,
$\text{op\_count}(g) \geq \tau_{\text{contract}}$, and per-element
feedback predicates entailing the target Bloom's level--using rule evaluation without LLM inference,
ensuring constant cost and formal verifiability. Failed scenes
trigger a bounded \texttt{re-Send} loop (max~2).

\paragraph{Assembly and Schema Compliance.}
Phase~5 instantiates the selected template as a React component via
the \textbf{Blueprint Assembler} and injects validated
content---the same orchestration layer produces all 15 mechanics
through component swapping, not code regeneration. QG4 performs
final blueprint validation with \texttt{is\_degraded} flag
tracking; inter-agent communication achieves 98.3\% Pydantic schema compliance\footnote{3.4 of every 200 inter-agent messages failed schema validation, measured across all pipeline runs.} \citep{wu2023,hong2023}.

\subsubsection{Deployment and Game Library}
The orchestration layer is model-agnostic: a declarative preset system
enables per-agent model selection across closed-source APIs
(GPT-4~\citealp{openai2023}, Gemini~\citealp{google2023}) and
open-source models (Llama~3~\citealp{grattafiori2024},
Mistral~\citealp{jiang2023}, Qwen~\citealp{qwen2025}) without
pipeline modification. The 50-game library (curated from the
evaluation corpus) serves as both demo set and regression corpus;
every game emits structured outcome data including score, interaction
trace, and inferred Bloom's level.

\subsection{Modular Game Engine}

The frontend implements a \textbf{plugin architecture}
(Figure~\ref{fig:system_flow}c): each of the 15 mechanics is a
self-contained React component registered by contract type, enabling
extension through registration without modifying orchestration layers.
Both template families share interaction primitives built on dnd-kit
\citep{dndkit2024} with custom collision detection and keyboard/touch
support. State management follows a dual architecture: Diagram Games
use a centralised Zustand store for multi-mechanic coordination;
Algorithm Games use localised reducer hooks for step-through
interactions. The engine is designed for WCAG-aligned keyboard
navigation and screen reader announcements; formal accessibility audit
is planned as part of the classroom evaluation phase.

\subsection{Pipeline Observability}

The demonstration includes a real-time observability dashboard
(Figure~\ref{fig:system_flow}b) with \textbf{three view modes}:
timeline, DAG graph (ReactFlow with execution-state highlighting), and
cluster view grouped by phase. Per-agent \textbf{token and cost
analytics} show stage-level consumption with USD breakdown.

\subsection{Design Validation}
Section~\ref{sec:eval} presents the full evaluation across all three architectures ($N = 200$, all 15 mechanics).
\section{Evaluation}
\label{sec:eval}

\subsection{Scope}
This evaluation measures \textbf{architectural validity}: validation
pass rate, token efficiency, and structural Bloom's alignment. It does
not measure learning outcome gains (see Limitations).
\subsection{Setup}
200 questions from five domains (biology, history, CS, mathematics,
linguistics) stratified across Bloom's levels and covering all 15
mechanics. All architectures used \textbf{GPT-4-turbo-2024-04-09}
\citep{openai2023} (temp.\ 0.3, seed 42) for planning/validation and
\textbf{gemini-1.5-pro-001} \citep{google2023} (temp.\ 0.4) for
asset generation, logged via LangSmith with per-call granularity.

\subsection{Baselines}
Five categories: \textbf{manual authoring} by five educators via
Genially/H5P (human-quality ceiling); \textbf{EdTech platforms}
(Kahoot, Quizlet, Nearpod, H5P) and GameGPT
\citep{chen2023gamegpt}; \textbf{Claude Code} under four prompting
conditions (zero-shot, one-shot content, one-shot instructional,
multi-turn) across 30 stratified questions each; and
\textbf{internal baselines} (Sequential Pipeline, ReAct Agent) on
all 200 questions covering both template families.
\subsection{Human rating methodology}
Educational Correctness and Playability ratings
(Table~\ref{tab:per_mechanic}) were collected from \textbf{five
domain-expert raters} (3 educators with $\geq$5 years experience; 2
SMEs per domain) using a behaviorally-anchored rubric. Raters were
\textbf{blind to system condition}. Inter-rater reliability was
acceptable (Educational Correctness: ICC$_{(2,5)}$ = 0.81, 95\% CI
[0.74, 0.87]; Playability: ICC$_{(2,5)}$ = 0.78, 95\% CI [0.71,
0.84]). The comparison between \gamedai{} (4.2/5) and manual authoring
(4.3/5) is not statistically significant ($t(199) = 1.04$,
$p = 0.30$). We note this result reflects insufficient statistical
power to detect a difference rather than confirmed equivalence;
formal equivalence testing with a pre-specified margin is
left for future work.

\subsection{Validation pass rate}
\gamedai{} achieves a VPR of \textbf{90.0\%}---17.5 percentage points
above ReAct Agents (72.5\%) and 33.3 points above the Sequential
Pipeline (56.7\%), confirmed significant ($\chi^2(2, N=600) = 57.0$,
$p < 0.001$, Cramér's $V = 0.31$). Architectural Validation Pass Rate (VPR) measures structural contract
compliance against \gamedai{}'s own FOL-based validators; it is an
internal architectural metric, not an independent measure of
pedagogical effectiveness.
\paragraph{Token consumption and cost.}
The 73\% token reduction from ReAct Agents to the DAG
(${\sim}$73,500 $\rightarrow$ ${\sim}$19,900 tokens/game) is
structural: architecture explains 87\% of token consumption variance
($\eta^2 = 0.87$, $F(2,\,597) = 1{,}996$, $p < 0.001$). We note
that ReAct agents perform self-correction loops by design; this
comparison partly reflects differing amounts of work done per
generation, not only efficiency. \gamedai{} is the only architecture
meeting the sub-\$0.50 cost requirement (Interactive Diagram Games
average \$0.46; Algorithm Games average \$0.43 due to fewer vision
model calls).

\subsection{Per-mechanic performance}
\begin{figure}[htbp]
    \centering
    \includegraphics[width=\linewidth]{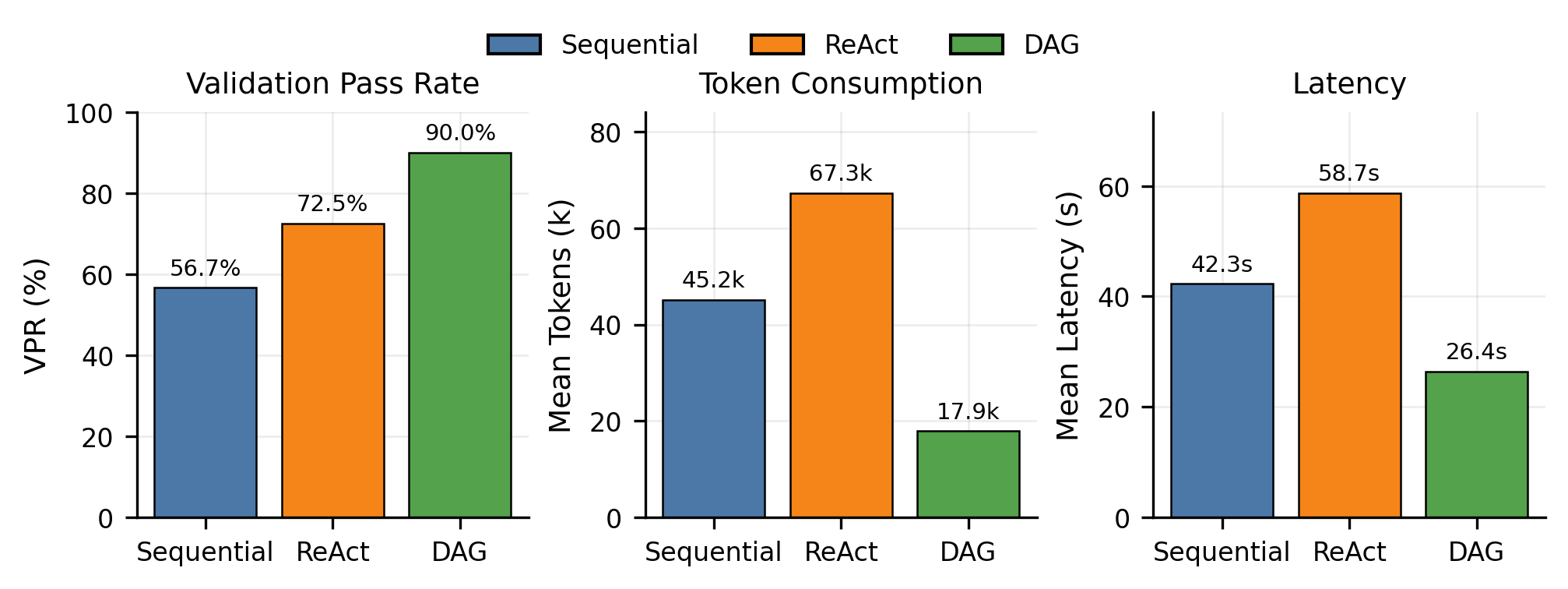}
    \caption{\textbf{Quality and efficiency metrics by architecture
    ($N = 200$ per condition, 15 mechanics).} Architecture explains
    87\% of token consumption variance ($\eta^2 = 0.87$); VPR gain
    of 17.5~pp over the next-best design.}
    \label{fig:arch_comparison}
\end{figure}

Figure~\ref{fig:arch_comparison} summarises results by architecture.
Across all 15 mechanics, VPR ranges from 96.2\%
(\texttt{DRAG\_DROP}) to 60.0\% (\texttt{DESC\_MATCHING}) for
Interactive Diagram Games, and from 94.4\%
(\texttt{STATE\_TRACER}) to 80.0\% (\texttt{CONSTRAINT\_PUZZLE}) for
Interactive Algorithm Games; mean educational correctness is 4.2/5.
Algorithm Games average higher token consumption
(${\sim}$23,500 vs.\ ${\sim}$17,900 tokens/game) but lower per-game
cost due to fewer vision model calls. Full per-mechanic breakdowns
are in Table~\ref{tab:per_mechanic}
(Appendix~\ref{app:per_mechanic}).

\section{Baselines Comparison}
\label{sec:baselines}

Table~\ref{tab:platform_comparison} summarizes all systems across
time, interactivity, cost, and Bloom's alignment; two findings
stand out.

\begin{table}[h]
\centering
\small
\setlength{\tabcolsep}{4pt}
\begin{tabularx}{\columnwidth}{@{} X l l l @{}}
\toprule
\textbf{System} 
  & \textbf{Time} 
  & \textbf{Cost} 
  & \textbf{Game Gen.} \\
\midrule
Kahoot \citep{wang2020}
  & 15--30 min 
  & \$7/mo\textsuperscript{a} 
  & Pre-built \\
Quizlet \citep{quizlet2024}
  & 20--40 min 
  & \$8/mo\textsuperscript{a} 
  & Pre-built \\
Genially \citep{genially2024}
  & 60--120 min 
  & \$25/mo\textsuperscript{a} 
  & Template \\
H5P \citep{h5p2024}
  & 40--90 min 
  & Free (OSS)
  & Template \\
GameGPT \citep{chen2023gamegpt}
  & ${\sim}$10 min 
  & \$0.60 
  & Template \\
Manual \citep{chapman2010}
  & 60--240 min 
  & \$50--150\textsuperscript{b}
  & Authored \\
\midrule
\textbf{\gamedai{} (DAG)}
  & \textbf{${<}$1 min} 
  & \textbf{\$0.46} 
  & \textbf{Dynamic} \\
\bottomrule
\end{tabularx}
\caption{Platform comparison of authoring latency and unit costs. \gamedai{} achieves sub-minute generation at \$0.46/game, outperforming both template-based AI and manual labor. \textsuperscript{a}Subscription fees for platform access; \textsuperscript{b}Estimated expert labor cost per unit.}
\label{tab:platform_comparison}
\end{table}
\gamedai{} compresses 60--240 minutes of expert authoring into under 
60 seconds at a fixed cost below the cheapest subscription tier of 
any listed platform. Expert raters scored it at 4.2/5 vs.\ 4.3/5 for 
manual authoring ($t(199) = 1.04$, $p = 0.30$); this non-significant 
result reflects insufficient statistical power ($n{=}5$), not 
confirmed equivalence---formal equivalence testing with a 
pre-specified margin is left for future work. GameGPT 
\citep{chen2023gamegpt} addresses creation speed but provides neither 
Bloom's targeting nor contract validation; MetaGPT \citep{hong2023} 
and AutoGen \citep{wu2023} provide role-bounded multi-agent 
coordination for software engineering tasks, with schemas governing 
code artifacts rather than pedagogical alignment predicates.

The Claude Code comparison functions as an \textbf{ablation of the
  contract mechanism}: Claude Code received identical learning
  objectives but not the mechanic contract schemas. Under these
  conditions, Claude Code produced functional games in 100\% of
  attempts---but only 23\% passed Bloom's alignment at zero-shot; the
  multi-turn ceiling of 67\% at \$0.80/run remains 23~pp below
  \gamedai{}'s 90\% VPR at lower cost
  (Table~\ref{tab:prompting_comparison}). This gap suggests FOL-based
  validation and phase-bounded generation provide structural guarantees
  beyond prompting alone \citep{ji2023,mislevy2003}; a controlled
  ablation providing equivalent schemas remains future work.

\begin{table}[ht]
\centering
\small
\setlength{\tabcolsep}{4pt}
\begin{tabularx}{\columnwidth}{@{}X
    S[table-format=2.1]
    S[table-format=3.1]
    S[table-format=1.2]@{}}
\toprule
\textbf{Prompting Strategy}
    & {\textbf{Align.}}
    & {\textbf{Tokens}}
    & {\textbf{Cost}} \\
    & {(\%)}
    & {(K)}
    & {(\$)} \\
\midrule
Zero-Shot              & 23.0 &  46.9 & 0.92 \\
One-Shot (Exemplar)     & 41.0 &  67.2 & 1.34 \\
One-Shot (Instr.)      & 48.0 &  74.4 & 1.51 \\
Multi-Turn             & 67.0 & 140.5 & 6.35 \\
\midrule
\textbf{\gamedai{} (DAG)}
    & \textbf{90.0} & \textbf{19.9} & \textbf{0.46} \\
\bottomrule
\end{tabularx}
\caption{Bloom's alignment, total token consumption, and total cost across
Claude Code prompting conditions and \gamedai{}. All Claude Code
runs use Claude-Opus-4-6; \gamedai{} uses
the full DAG pipeline.}
\label{tab:prompting_comparison}
\end{table}

\section{Conclusion and Future Work}

We present \gamedai{}, a hierarchical multi-agent framework that
generates Bloom's-aligned educational games through formal mechanic
contracts and deterministic Quality Gates---achieving 90\% validation
pass rate, 73\% token reduction over ReAct baselines, and sub-minute
generation at \$0.46/game. Within this configuration, phase-bounded architecture with contract validation correlates with improved alignment and token efficiency; replication across model
families remains open. Future work targets human-in-the-loop blueprint
negotiation, frame-based game engines (Phaser.js) for physics
mechanics, expanded template families, and large-scale classroom
evaluation measuring learning outcome gains.
\section*{Limitations}
\noindent\textbf{Schema coverage.}
Spatial anchoring for \texttt{DESC\_MATCHING} and \texttt{TRACE\_PATH}
is underspecified, producing 14 of 20 validation failures;
relational-link extensions are underway.

\noindent\textbf{Model and language scope.}
Metrics reflect a GPT-4-turbo/Gemini configuration; open-source
substitution (Llama~3, Mistral) is supported but not benchmarked.
Games are English-only; multilingual support is planned.

\noindent\textbf{Student-facing validation.}
The evaluation measures architectural validity, not learning outcomes.
Expert ratings ($n{=}5$) may diverge from learner judgements;
classroom studies are the primary future direction.

\section*{Broader Impact}
\gamedai{} lowers cost and expertise barriers to structurally
validated game authoring. Current evidence is limited to architectural
metrics and expert ratings; the system should be understood as a
validated authoring tool, not a proven pedagogical intervention.
Open-source release and model-agnostic deployment support
reproducibility and data sovereignty.

\noindent\textbf{Risks.}
Quality Gates validate structure but not factual accuracy---incorrect
domain knowledge can propagate into games. Mitigations include curated
knowledge retrieval, deterministic validators, and an instructor-facing
observability dashboard. \gamedai{} is designed to augment, not
replace, pedagogical judgement.

\bibliography{gamedai}

\appendix
\section{Bloom's Mapping and Mechanic Contracts}
\label{app:blooms_mapping}
\label{app:mechanic_specs}

Table~\ref{tab:blooms_mapping} maps Bloom's levels to mechanics
grounded in \citet{anderson2001} and the LM-GM framework
\citep{arnab2015}; full Pydantic contracts are in the repository.

\begin{table}[ht]
\centering
\footnotesize
\setlength{\tabcolsep}{3pt}
\begin{tabular}{@{}l l l@{}}
\toprule
\textbf{Bloom's}
  & \textbf{Fam.}
  & \textbf{Mechanics} \\
\midrule
\textit{Remember}
  & ID   & Click-to-Id., Memory Match \\
\textit{Understand}
  & ID   & Drag-Drop, Desc.\ Match \\
\textit{Apply}
  & Both & Trace Path, Seq., State Tr.\textsuperscript{A} \\
\textit{Analyze}
  & Both & Sorting, Hier., Bug H.\textsuperscript{A},
           Cmplx.\textsuperscript{A} \\
\textit{Evaluate}
  & ID   & Compare, Branching \\
\textit{Create}
  & Algo & Algo.\ Builder\textsuperscript{A},
           Constr.\ Pzl.\textsuperscript{A} \\
\bottomrule
\end{tabular}
\caption{Bloom's-to-mechanic mapping.
Fam.: ID = Interactive Diagram, Algo = Algorithm, Both = shared;
\textsuperscript{A} = Algorithm Game mechanic.}
\label{tab:blooms_mapping}
\end{table}
\section{Per-Mechanic Results}
\label{app:per_mechanic}

Table~\ref{tab:per_mechanic} disaggregates VPR, token consumption,
latency, and human ratings across all 15 mechanics. Fully constrained
schemas achieve $\geq$90\% VPR; the two lowest-performing mechanics
(\texttt{DESC\_MATCH}, \texttt{CONSTR\_PUZZLE}) share root causes
detailed in Appendix~\ref{app:failure}. Mechanics with $N \leq 10$
should be interpreted with caution; per-mechanic 95\% Wilson
confidence intervals are available in the repository.

\begin{table}[ht]
\centering
\small
\setlength{\tabcolsep}{3pt}
\begin{tabular}{
    l
    c
    S[table-format=2.1]
    S[table-format=2.1]
    S[table-format=2.1]
    S[table-format=1.1]
    S[table-format=2.1]
}
\toprule
\textbf{Mechanic}
  & \textbf{N}
  & {\textbf{VPR}}
  & {\textbf{Tok.}}
  & {\textbf{Lat.}}
  & {\textbf{Edu.}}
  & {\textbf{Play}} \\
  & & {(\%)} & {(K)} & {(s)} & {(1--5)} & {(\%)} \\
\midrule
\multicolumn{7}{l}{\textit{Interactive Diagram Games}} \\
\texttt{DRAG\_DROP}     & 26 & 96.2 & 18.2 & 27.0 & 4.4 & 96.2 \\
\texttt{SEQUENCING}     & 16 & 93.8 & 17.0 & 25.0 & 4.5 & 93.8 \\
\texttt{CLICK\_TO\_ID}  & 14 & 92.9 & 16.8 & 24.0 & 4.3 & 92.9 \\
\texttt{SORTING}        & 12 & 91.7 & 18.5 & 28.0 & 4.2 & 91.7 \\
\texttt{MEMORY\_MATCH}  & 12 & 91.7 & 16.2 & 23.0 & 4.3 & 91.7 \\
\texttt{BRANCHING}      & 10 & 90.0 & 19.5 & 30.0 & 4.1 & 90.0 \\
\texttt{COMPARE}        &  8 & 87.5 & 20.1 & 31.0 & 4.0 & 87.5 \\
\texttt{HIERARCHICAL}   &  8 & 87.5 & 22.4 & 35.0 & 3.9 & 87.5 \\
\texttt{TRACE\_PATH}    & 14 & 85.7 & 17.5 & 26.0 & 4.1 & 85.7 \\
\texttt{DESC\_MATCH}    & 10 & 60.0 & 15.8 & 22.0 & 3.8 & 75.0 \\
\midrule
\multicolumn{7}{l}{\textit{Interactive Algorithm Games}} \\
\texttt{STATE\_TRACER}  & 18 & 94.4 & 21.3 & 32.0 & 4.4 & 94.4 \\
\texttt{BUG\_HUNTER}    & 16 & 93.8 & 23.8 & 36.0 & 4.2 & 87.5 \\
\texttt{ALGO\_BUILDER}  & 14 & 92.9 & 25.2 & 38.0 & 4.3 & 92.9 \\
\texttt{COMPLEXITY}     & 12 & 91.7 & 22.7 & 34.0 & 4.1 & 83.3 \\
\texttt{CONSTR\_PUZZLE} & 10 & 80.0 & 26.5 & 40.0 & 3.9 & 80.0 \\
\midrule
\textbf{Overall}
  & \textbf{200} & \textbf{90.0} & \textbf{19.9}
  & \textbf{29.8} & \textbf{4.2} & \textbf{89.8} \\
\bottomrule
\end{tabular}
\caption{Per-mechanic metrics ($N = 200$, 15 mechanics). Edu./Play:
mean human ratings from 5 blinded raters (ICC $> 0.78$). Algorithm
Games average higher token consumption but lower per-game cost (no
vision model calls). Mechanics with $N \leq 10$: interpret with
caution; Wilson 95\% CIs in repository.}
\label{tab:per_mechanic}
\end{table}

\section{Failure Taxonomy}
\label{app:failure}

Of the 20 DAG failures across 200 questions, 14 occur in Interactive
Diagram mechanics and 6 in Algorithm Games. The dominant root cause
is \textbf{schema underspecification}, not LLM hallucination:
generated content is factually correct but lacks structural fields
required by the FOL-based contract validator. All failure types are
tractable schema engineering problems currently under active
remediation.

\begin{table}[ht]
\centering
\footnotesize
\setlength{\tabcolsep}{3pt}
\begin{tabularx}{\columnwidth}{@{}l c l X@{}}
\toprule
\textbf{Mechanic} & \textbf{N} & \textbf{Gate}
  & \textbf{Error / Root Cause} \\
\midrule
\multicolumn{4}{l}{\textit{Interactive Diagram Games (14 failures)}} \\
\texttt{DESC\_MATCH}
  & 4 & QG3
  & \texttt{BLOOM\_OP\_COUNT\_FAIL}; pairs lack relational links \\
\texttt{TRACE\_PATH}
  & 2 & QG3
  & \texttt{ANCHOR\_OOB}; SVG coords outside bounding box \\
\texttt{COMPARE}
  & 1 & QG2
  & \texttt{ASSET\_SCHEMA\_MISMATCH}; axis label missing \\
\texttt{HIERARCHICAL}
  & 1 & QG3
  & \texttt{DEPTH\_MISMATCH}; tree depth $<$ contract min \\
6 other ID mech.
  & 6 & QG2/3
  & Region overlap (2); state/schema violations (4) \\
\midrule
\multicolumn{4}{l}{\textit{Interactive Algorithm Games (6 failures)}} \\
\texttt{CONSTR\_PZL}
  & 2 & QG3
  & \texttt{CONSTRAINT\_UNSAT}; FOL rules form unsat.\ set \\
\texttt{COMPLEXITY}
  & 1 & QG3
  & \texttt{CLASS\_MISMATCH}; generated $\neq$ target class \\
3 other Algo mech.
  & 3 & QG2/3
  & Placement, ordering, state transition errors \\
\bottomrule
\end{tabularx}
\caption{Failure taxonomy (20 failures, $N = 200$, 15 mechanics).
All attributable to schema underspecification, not LLM hallucination.
FOL-based validators catch structural violations deterministically.}
\label{tab:failure_analysis}
\end{table}


\definecolor{layerblue}{RGB}{210,228,248}
\definecolor{layergreen}{RGB}{210,238,210}
\definecolor{layeryellow}{RGB}{255,243,205}
\definecolor{layerpink}{RGB}{248,215,218}
\definecolor{layerpurple}{RGB}{235,220,248}

\tcbset{
  promptbox/.style={
    enhanced,
    breakable,
    arc=4pt,
    boxrule=0.6pt,
    left=6pt, right=6pt,
    top=4pt, bottom=4pt,
    fontupper=\ttfamily\footnotesize,
    attach boxed title to top left={
      yshift=-2mm, xshift=4mm
    },
    boxed title style={
      arc=3pt,
      boxrule=0.5pt,
      left=4pt, right=4pt,
      top=1pt, bottom=1pt,
      fontupper=\itshape\bfseries\small
    }
  }
}

\section{Mechanic-Specific Agent Prompting}                                                                                                                                                                                                                                                        
  \label{app:prompt_examples}                                                                                                                                                                                                                                                                      
                                                                                                                                                                                                                                                                                                     
  \begin{tcolorbox}[                                                                                                                                                                                                                                                                                 
    enhanced, breakable,                                                                                                                                                                                                                                                                             
    arc=0pt, boxrule=1pt,                                                                 
    colback=white, colframe=black!80,       
    title={\textbf{Game Concept Designer (Phase~1)} \hfill
           \small\textit{DRAG\_DROP · Analyze Level}},
    fonttitle=\small\bfseries\color{white},
    left=0pt, right=0pt, top=0pt, bottom=0pt,
    toptitle=3pt, bottomtitle=3pt
  ]

  \begin{tcolorbox}[
    colback=blue!10, colframe=blue!30,
    boxrule=0pt, arc=0pt,
    title={\textit{System Role}}, fonttitle=\small\itshape\color{black},
    left=6pt, right=6pt, top=3pt, bottom=3pt
  ]
  \footnotesize
  You are an expert educational game designer. Transform learning questions into GameConcept JSON: title, narrative, scenes, mechanics, and WHY each mechanic matches the learning objective. Focus on WHAT and WHY---visual design comes later.
  \end{tcolorbox}

  \begin{tcolorbox}[
    colback=green!10, colframe=green!30,
    boxrule=0pt, arc=0pt,
    title={\textit{Design Principles}}, fonttitle=\small\itshape\color{black},
    left=6pt, right=6pt, top=3pt, bottom=3pt
  ]
  \footnotesize
  Match mechanic to objective: \texttt{drag\_drop}$=$spatial, \texttt{trace\_path}$=$process, \texttt{sequencing}$=$temporal, \texttt{sorting}$=$classification.
  Scenes can chain 1--3 mechanics via \texttt{advance\_trigger}. Zone-based mechanics share a diagram; content-only (\texttt{sequencing}, \texttt{memory\_match}) set \texttt{needs\_diagram=false}.
  \end{tcolorbox}

  \begin{tcolorbox}[
    colback=yellow!15, colframe=yellow!35,
    boxrule=0pt, arc=0pt,
    title={\textit{Task Prompt (assembled from state)}}, fonttitle=\small\itshape\color{black},
    left=6pt, right=6pt, top=3pt, bottom=3pt
  ]
  \footnotesize
  \texttt{\#\# Question:} Label the parts of a plant cell.\\
  \texttt{\#\# Context:} Bloom's: analyze · Biology · intermediate\\
  \texttt{\#\# Domain Knowledge:} Labels: Chloroplast, Mitochondria, Cell Wall, Vacuole, Nucleus, Ribosome. Descriptions: \{Chloroplast: ``Conducts photosynthesis\ldots''\}. Needs: labels=true, sequence=false.\\
  \texttt{\#\# Capabilities:} \texttt{[drag\_drop, click\_to\_identify, trace\_path, sequencing, sorting, description\_matching, memory\_match, branching]}
  \end{tcolorbox}

  \begin{tcolorbox}[
    colback=purple!10, colframe=purple!30,
    boxrule=0pt, arc=0pt,
    title={\textit{Output (GameConcept, Pydantic)}}, fonttitle=\small\itshape\color{black},
    left=6pt, right=6pt, top=3pt, bottom=3pt
  ]
  \footnotesize
  \texttt{\{title, subject, difficulty, narrative\_theme, all\_zone\_labels[], distractor\_labels[], scenes[\{title, learning\_goal, zone\_labels[], needs\_diagram, mechanics[\{mechanic\_type, learning\_purpose, expected\_item\_count, advance\_trigger\}]\}]\}}
  \end{tcolorbox}

  \begin{tcolorbox}[
    colback=red!10, colframe=red!30,
    boxrule=0pt, arc=0pt,
    title={\textit{QG1 Retry Injection}},
    fonttitle=\small\itshape\color{black},
    left=6pt, right=6pt, top=3pt, bottom=3pt
  ]
  \footnotesize
  \texttt{RETRY:} scene\_1 uses \texttt{sequencing} but no sequence data available; \texttt{distractor\_labels} overlaps \texttt{all\_zone\_labels}; missing \texttt{estimated\_duration\_minutes}.
  \end{tcolorbox}

  \end{tcolorbox}\section{Architectural Evolution: Pipeline Diagrams}
\label{app:evolution}

\begin{figure}[ht]
    \centering
    \includegraphics[width=\columnwidth]{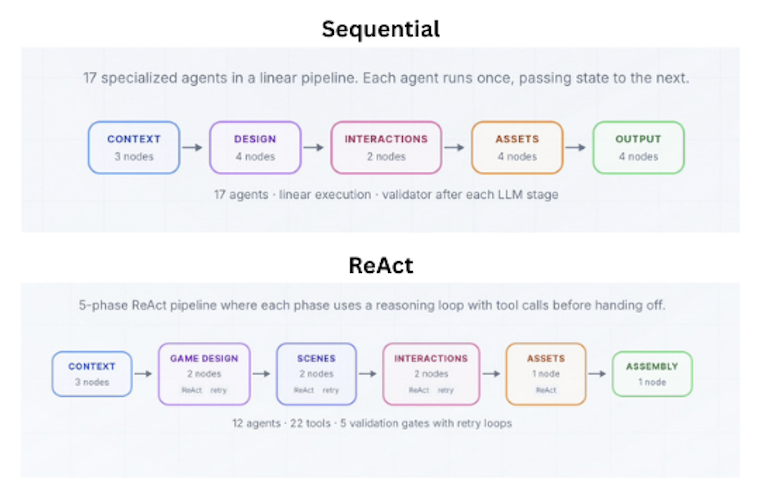}
    \caption{\textbf{Pipeline Evolution: Sequential and ReAct
Architectures.} The Sequential design runs 17 specialised agents
linearly with a validator after each LLM stage. The ReAct design uses
a 5-phase reasoning loop with 12 agents, 22 tools, and 5 validation
gates with retry loops.}
    \label{fig:agent_pipelines}
\end{figure}

\end{document}